\documentclass[lettersize,journal]{IEEEtran}
\usepackage{amsmath,amsfonts}
\usepackage{algorithmic}
\usepackage{algorithm}
\usepackage{array}
\usepackage[caption=false,font=normalsize,labelfont=sf,textfont=sf]{subfig}
\usepackage{textcomp}
\usepackage{stfloats}
\usepackage{url}
\usepackage{verbatim}
\usepackage{graphicx}
\usepackage{mathtools}
\usepackage{booktabs}
\usepackage{multicol,multirow}
\usepackage{enumitem}
\usepackage{makecell}
\usepackage{cite}
\usepackage{hyperref}

\usepackage{listings}
\usepackage{xcolor}

\definecolor{codegreen}{rgb}{0,0.6,0}
\definecolor{codegray}{rgb}{0.5,0.5,0.5}
\definecolor{codepurple}{rgb}{0.58,0,0.82}
\definecolor{backcolour}{rgb}{0.95,0.95,0.92}
\lstdefinestyle{mystyle}{
    backgroundcolor=\color{backcolour},   
    commentstyle=\color{codegreen},
    keywordstyle=\color{magenta},
    numberstyle=\tiny\color{codegray},
    stringstyle=\color{codepurple},
    basicstyle=\ttfamily\footnotesize,
    breakatwhitespace=false,         
    breaklines=true,                 
    captionpos=none,                    
    keepspaces=true,                 
    numbers=none,                    
    numbersep=5pt,                  
    showspaces=false,                
    showstringspaces=false,
    showtabs=false,                  
    tabsize=2
}

\newcommand{\ie}{\textit{i.e.}}

\newcommand{\etal}{\textit{et al.}}
\DeclarePairedDelimiter{\norm}{\lVert}{\rVert}
\begin{document}

\title{Delving Deep into Semantic Relation Distillation}
%\title{Delving Deep into Semantic Relation Modeling for Knowledge Distillation}

\author{Zhaoyi~Yan,
        Kangjun~Liu,
        Qixiang~Ye
        % <-this % stops a space

%\thanks{The submitted manuscript is xx.}
\thanks{Q. Ye is with the School of Electronic, Electrical and Communication Engineering, University of Chinese Academy of Sciences, Beijing, China and also with Peng Cheng Laboratory, Shenzhen, China (e-mail: qxye@ucas.ac.cn).}

\thanks{Z. Yan is with School of with the School of Computer Science and Technology, Harbin Institute of Technology, Harbin 150001, China (e-mail: yanzhaoyi@outlook.com).}

\thanks{K. Liu is with Peng Cheng Laboratory, Shenzhen, China (e-mail:  liukj@pcl.ac.cn). }

\thanks{Corresponding Author: Qixiang Ye.}
}

\markboth{Journal of \LaTeX\ Class Files,~Vol.~14, No.~8, August~2021}%
{Shell \MakeLowercase{\textit{et al.}}: A Sample Article Using IEEEtran.cls for IEEE Journals}


\maketitle

\begin{abstract}
Knowledge distillation has become a cornerstone technique in deep learning, facilitating the transfer of knowledge from complex models to lightweight counterparts. 
Traditional distillation approaches focus on transferring knowledge at the instance level, but fail to capture nuanced semantic relationships within the data. 
In response, this paper introduces a novel methodology, Semantics-based Relation Knowledge Distillation (SeRKD), which reimagines knowledge distillation through a semantics-relation lens among each sample. 
By leveraging semantic components, \ie, superpixels, SeRKD enables a more comprehensive and context-aware transfer of knowledge, which skillfully integrates superpixel-based semantic extraction with relation-based knowledge distillation for a sophisticated model compression and distillation.
Particularly, the proposed method is naturally relevant in the domain of Vision Transformers (ViTs), where visual tokens serve as fundamental units of representation. 
Experimental evaluations on benchmark datasets demonstrate the superiority of SeRKD over existing methods, underscoring its efficacy in enhancing model performance and generalization capabilities. 
%
% Overall, this paper presents a paradigm-shifting approach to knowledge distillation, opening new avenues for more effective and contextually rich knowledge transfer in machine learning models.
\end{abstract}


\section{Introduction}
\label{sec:intro}

In the past decade, knowledge distillation~\cite{hinton2015distilling} has become a cornerstone technique in deep learning, especially in scenarios where deploying large-scale models is impractical due to computational constraints or deployment considerations. This method enables the transfer of knowledge from complex, computationally expensive models, referred to as ``teacher" models, to more lightweight ``student" models. Its applications span various domains, including computer vision~\cite{gou2021knowledge}, natural language processing~\cite{sun2019patient}, and speech recognition~\cite{yoon2021tutornet}, where model efficiency and scalability are paramount.

Traditionally, knowledge distillation methods have always focused predominantly on transferring information at the instance level. These methods can be categorized into logit-based approaches~\cite{li2023boosting, beyer2022knowledge, zhao2022decoupled, yuan2020revisiting, xu2023improving} and feature-based approaches~\cite{fitnet, heo2019comprehensive, chen2021distilling, li2023rethinking, zhang2021student}, which aim to match the output probabilities or intermediate representations between teacher and student models. Additionally, relation-based approaches~\cite{Park_2019_CVPR, huang2022knowledge} have gained attention for capturing complex dependencies and intrinsic knowledge by modeling relationships or correlations between different instances or categories.

However, these methods often fail to capture the intricate semantic relationships inherent in the data. This limitation has spurred research into methodologies that leverage semantic understanding and contextual information to enhance the distillation process. Meanwhile, recent advancements~\cite{jin2023multi, lin2022knowledge} underscore the necessity of moving beyond instance-level knowledge transfer, emphasizing the importance of capturing and transferring more nuanced and contextually relevant information. This shift is particularly crucial given the increasing availability of large-scale datasets and the development of sophisticated model architectures.

\begin{figure*}[t]
    \centering
    \includegraphics[width=\linewidth]{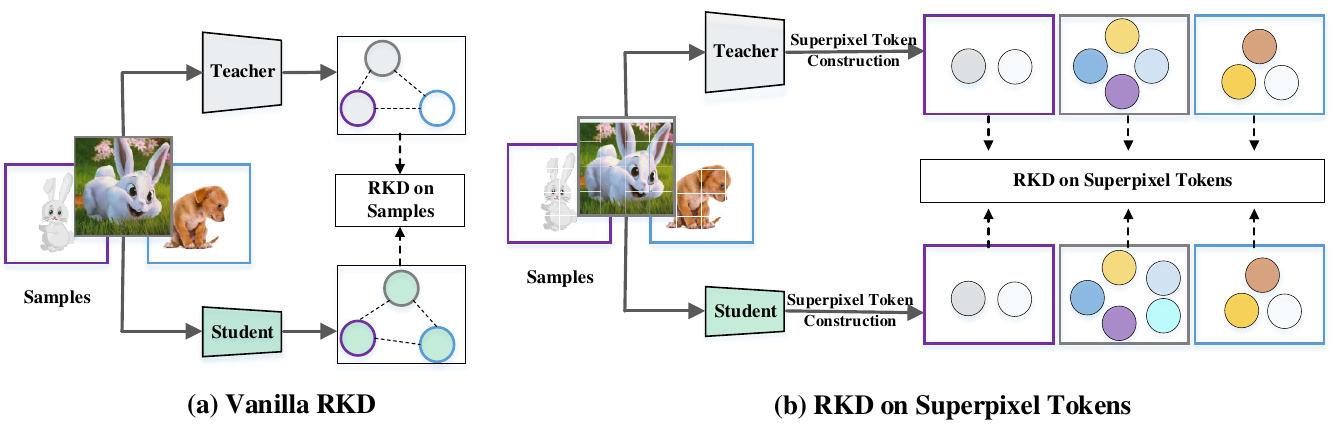}
    \caption{Comparison of different relation-based distillation techniques. While vanilla RKD focuses on building relationships among samples, our method distills relational knowledge among semantic-superpixel tokens at an instance level.}
    \label{fig:intro}
\end{figure*}

Inspired by these observations, we propose a novel method called Semantics-based Relation Knowledge Distillation (SeRKD). This approach leverages superpixels~\cite{jampani2018superpixel, huang2023vision} to extract semantic components, facilitating a more comprehensive and context-aware transfer of knowledge. By combining superpixel-based semantic extraction with relation-based knowledge distillation, SeRKD offers a sophisticated and nuanced approach to model compression and distillation. Our methodology reimagines knowledge distillation from a semantics-centric perspective, providing more effective knowledge transfer compared to conventional instance-level methods. Building upon the foundational work of Hinton et al.~\cite{hinton2015distilling}, our approach goes beyond simply matching output probabilities or intermediate representations by incorporating semantic relationships within the data.

As illustrated in Figure~\ref{fig:intro}, our approach differs from traditional relation-based distillation techniques by focusing on semantic-superpixel tokens at the instance level. The core of our methodology lies in the extraction and utilization of semantic relationships within the data, capturing relationships between different semantic components to enrich the distillation process with valuable contextual insights. This semantics-centric approach enables a more comprehensive and contextually aware transfer of knowledge, leading to improved model performance and generalization capabilities.

Our methodology is particularly well-suited for Vision Transformers (ViTs), which rely on visual tokens as fundamental units of representation. Aligned with the token-based architecture of ViTs, our method ensures seamless integration and enhances performance. By extracting knowledge from semantic parts represented by superpixels, our approach provides a fine-grained understanding of image content. This detailed semantic extraction significantly boosts the efficiency and accuracy of the distillation process.

The contributions of this study are as follows:
\begin{itemize}
\item We introduce a semantics-centric approach to knowledge distillation, providing a deeper and more nuanced understanding of data and enabling more effective knowledge transfer compared to conventional instance-level methods.
\item Technically, our proposed framework, Semantics-based Relation Knowledge Distillation (SeRKD), combines superpixel extraction with relation-based knowledge distillation, offering a sophisticated and contextually rich approach to model compression and distillation.
\item Experimental evaluations conducted on various benchmark datasets demonstrate the superior performance of SeRKD over existing methods, highlighting its efficacy in enhancing model performance and generalization capabilities.
\end{itemize}

In summary, this paper presents a novel approach to knowledge distillation that transcends the limitations of traditional methods by embracing a semantics-centric perspective and integrating recent advancements in superpixel-based segmentation techniques. By enriching the distillation process with semantic understanding and contextual information, our methodology opens new avenues for more effective and contextually rich knowledge transfer in machine learning models.

\section{Related Work}
\label{sec:related_work}
In this section, we firstly briefly review some recent works regarding knowledge distillation. Secondly,  existing works about superpixel methods in model training are investigated.

\subsection{Knowledge Distillation}
Knowledge distillation, originating from the seminal work of Hinton et al. \cite{hinton2015distilling}, has evolved as a fundamental methodology for model training and compression. Numerous related works \cite{beyer2022knowledge, zhao2022decoupled, yuan2020revisiting, zhang2024collaborative} have subsequently emerged, focusing on utilizing the teacher's predictions as ``soft" labels to guide and supervise students. Beyond such logit-based methods, this avenue expands with various works \cite{fitnet, chen2021distilling, li2023rethinking}, delving into diverse perspectives of distilling feature-based knowledge, which usually transfer the spatial-wise knowledge in teacher model's intermediate features for adequate representation learning in student model.

One of the pivotal advancements in knowledge distillation is the introduction of Relation Knowledge Distillation (RKD) \cite{Park_2019_CVPR}. RKD stands out as a significant contribution, particularly for its emphasis on correlation learning within the same batch from one model to another. The method effectively captures nuanced relationships between data samples, contributing to improved knowledge transfer, which motivates us to explore further about the semantic correlation distillation.

However, a prevailing trend in the literature is the dominance of methods designed for CNNs, with limited applicability to prevalent transformer-based models \cite{vaswani2017attention}. Notably, the Vision Transformer (ViT) series \cite{vit} has gained prominence as the most widely used network architecture. The scarcity of methods suitable for transformer-based models motivates the exploration of novel techniques that can cater to the evolving landscape of modern deep learning architectures. To contribute to the Vision Transformer (ViT) community through Knowledge Distillation (KD), DeiT \cite{deit} firstly suggests distilling knowledge from CNNs to Vision Transformers. 
Subsequently, DearKD\cite{chen2022dearkd} introduces a dual-stage learning framework, where knowledge distillation occurs exclusively from the intermediate features of CNNs during the initial stage. 
CSKD \cite{zhao2023cumulative} employs a technique to distill spatial-wise knowledge to all patch tokens of ViT directly from the corresponding spatial responses of CNNs, eliminating the need for utilizing any intermediate features. 
Recently, G2SD \cite{huang2024generic} enables effective bidirectional distillation between CNNs and ViTs through a generic-to-specific framework that transfers both task-agnostic and task-specific knowledge.
Those previous works prove the potential of distilling knowledge from CNNs to Vision Transformers, but lacking the exploration concerning the semantic distillation from well-learned Vision Transformers.

In departure from previous knowledge distillation methods, we propose the Semantics-based Relation Knowledge Distillation (SeRKD) algorithm. Distinguished by its part-wise correlation learning approach, SeRKD is uniquely positioned to enhance knowledge transfer learning for a diverse array of transformer-based architectures, including the widely adopted ViT series, while also accommodating CNN-based models. The versatility of SeRKD makes it a valuable contribution, addressing the limitations of existing methods and opening new avenues for effective knowledge distillation across various deep learning architectures.

\subsection{Superpixel Methods}

Superpixel algorithms are fundamentally categorized into graph-based and clustering-based methods. Graph-based methods conceptualize image pixels as nodes within a graph structure, segmenting these nodes based on the connectivity of adjacent pixel edges, as detailed in seminal works by ~\cite{ren2003learning, felzenszwalb2004efficient, liu2011entropy}. Conversely, clustering-based methods employ established clustering techniques like $k$-means to form superpixels, utilizing diverse feature representations to enhance the granularity of segmentation, as explored by~\cite{achanta2012slic, liu2016manifold}.
ETPS~\cite{yao2015real} and SEEDS~\cite{van2015seeds} initially partition the image into regular grids and utilize different energy functions to optimize the exchange of pixels between neighboring superpixels.

The advent of deep learning has precipitated a shift towards more sophisticated deep clustering approaches, as evidenced by the literature ~\cite{yeo2017superpixel, achanta2017superpixels, yang2020superpixel, cai2021revisiting}. These methods seek to harness deep feature representations to augment the efficiency and accuracy of superpixel generation. Notably, SEAL~\cite{tu2018learning} integrates deep learning features with traditional superpixel algorithms for enhanced feature learning. Similarly, SSN~\cite{jampani2018superpixel} introduces an innovative end-to-end differentiable superpixel segmentation framework.
Followed by SSN, SPIN~\cite{zhang2023lightweight} introduces Intra-Superpixel Attention (ISPA) and
Superpixel Cross Attention (SPCA) modules in ViTs for super-resolution tasks.
Similarly, STViT~\cite{huang2023vision} proposes a super token sampling algorithm to further refine the efficiency and effectiveness of superpixel segmentation.

In this paper, we construct relation structures for knowledge distillation using semantic parts derived from superpixel algorithms. Leveraging advanced methods like SSN \cite{jampani2018superpixel}, SPIN \cite{zhang2023lightweight}, or STViT \cite{huang2023vision}, we enhance the distillation process through meaningful image segmentation. This approach captures semantic relationships within the data, facilitating more effective knowledge transfer and improving performance across CNNs and Vision Transformers.

\section{Preliminary}
\subsection{Relation-based Knowledge Distillation}

Relation-based knowledge distillation aims to encapsulate the relational dynamics between training examples in the feature/output representation space. Such relation-based knowledge can capture the structure information in the data embedding space~\cite{huang2022knowledge, Park_2019_CVPR, Peng_2019_ICCV, Yang_2022_CVPR, Liu_2019_CVPR}.
Park~\etal~\cite{Park_2019_CVPR} exploit the distance-wise metric $\mathcal{L}_{\text{RD}}$ and angle-wise metric $\mathcal{L}_{\text{RA}}$ among training samples in the batch.
$\mathcal{L}_{\text{RD}}$ is defined over pairs of examples and is concerned with the Euclidean distances in the representation space. For a pair of examples, the distance-wise potential $\psi_{\text{D}}(u_i, u_j)$ is the normalized Euclidean distance between them:
\begin{align}
\psi_{\text{D}}(u_i, u_j) = \frac{1}{\nu} \norm{u_i - u_j}_2,
\end{align}
where $\nu$ is the normalization factor and $u_i$ is the output representation of the student/teacher network. The loss, $\mathcal{L}_{\text{RD}}$, then minimizes the Huber loss $l_{\delta}$ between the student's and teacher's distance-wise potentials, fostering a similar relational structure in the student.  Huber loss $l_{\delta}$ is formulated as:
\begin{align}
    \label{eq:huber_loss}
    l_{\delta} = \begin{cases}
     \frac{1}{2}{(x-y)^2}                  & \text{for } |x-y| \le 1, \\
     |x-y| - \frac{1}{2}, & \text{otherwise.}
    \end{cases}
\end{align}

Angle-wise Distillation Loss ($\mathcal{L}_{\text{RA}}$) focuses on the angles formed by triplets of examples, capturing higher-order relational information. The angle-wise potential $\psi_{\text{A}}(u_i, u_j, u_k)$ is the cosine of the angle formed by three data points:
\begin{align}
\psi_{\text{A}}(u_i, u_j, u_k) = \langle \mathbf{e}^{ij}, \mathbf{e}^{kj} \rangle,
\end{align}
where $\mathbf{e}^{ij}$ and $\mathbf{e}^{kj}$ are unit vectors. The corresponding loss, $\mathcal{L}_{\text{RA}}$, minimizes the Huber loss between the teacher's and student's angle-wise potentials.

Training with RKD involves combining these relational losses with a task-specific loss ($\mathcal{L}_\mathrm{task}$). The holistic objective is thus:
    \begin{align}
    \mathcal{L}_{\text{Total}} = \mathcal{L}_\mathrm{task} + \lambda_{D} \mathcal{L}_{\text{RD}} + \lambda_{A}\mathcal{L}_{\text{RA}},
    \end{align}
where $\lambda_{D}$ and $\lambda_{A}$ are the weights of the losses and all possible pairs and triplets from a given mini-batch are utilized for distillation.

These relational losses facilitate a more nuanced transfer of knowledge, focusing not just on direct outputs but also on the intricate geometric relationships between them, promising a richer and more flexible learning experience for the student model.

\subsection{Transformer Representations}
The Vision Transformer (ViT~\cite{vit}) framework transforms an input image into a structured sequence of vector representations, treating the image as a series of discrete `words' or patches. For a given image $I \in \mathcal{R}^{H \times W \times C}$, it is systematically segmented into $N = \frac{H \times W}{P^2}$ distinct patches $\{I_i^p\}_{i=1}^N$, with $H$, $W$, $C$, and $P$ denoting the height, width, number of channels, and patch dimension, respectively. Each patch $I_i^p$ represents a flattened vector of dimension $N \times (P^2C)$. This approach is demonstrated using a typical image size of $224 \times 224 \times 3$, which is segmented into a $14 \times 14$ patch grid, with each patch measuring $16 \times 16 \times 3$. Embedded positional encodings enhance the spatial relevance of these patches. Subsequent processing through the Transformer's architecture, which includes layers of multi-head self-attention~\cite{transformer} and feed-forward networks, refines these vectors into advanced image representations.

\subsection{Superpixel Sampling Networks}
\label{sec:snn}
Superpixel Sampling Networks (SSN~\cite{jampani2018superpixel}) introduced a differentiable approach for generating superpixels. Let $\boldsymbol{F}_{i}$ represent the feature value at pixel $i$ and $S_j$ denote superpixel $j$. The process begins by initializing superpixel centers using averaged features within regular grid cells, typically a $3\times3$ grid.
Superpixels are refined iteratively by updating the association between pixels and superpixels using a Radial Basis Function (RBF) kernel. The association matrix at iteration $t$, denoted as $Q^t \in \mathcal{R}^{n \times m}$, where $n$ is the total number of spatial pixels and $m$ is the number of superpixels, is updated as follows:
\begin{equation}
\label{eqn:q}
    Q_{ij}^t = \exp\left(-\|\boldsymbol{F}_i - S_j^{t-1}\|^2\right),
\end{equation}
reflecting the association based on the squared Euclidean distance between the feature of each pixel and the superpixel center from the previous iteration $(t-1)$.

To enhance computational efficiency, SSN restricts the computation of associations to the $9$ nearest neighboring pixels. The superpixel centers are then updated in each iteration by aggregating the associated features:
\begin{equation}
\label{eqn:s}
    S_j^t = \frac{1}{Z^{t}_{j}}\sum_{i} Q_{ij}^t \boldsymbol{F}_i,
\end{equation}
where $S_j^t$ denotes the updated superpixel center at current iteration $t$ and $Z^{t}_j=\sum_iQ^t_{ij}$.

\section{The Proposed Method}
\label{sec:method}

\subsection{Building Superpixel Tokens}
\label{sec:building_seman}

\subsubsection{Building Superpixel Tokens for ViTs}
Given an input image $I \in \mathcal{R}^{H \times W \times C}$, it is initially processed by the ViT to transform the image into a sequence of patch tokens $\{T_i\}_{i=1}^N$, where each token $\boldsymbol{T}_i$ corresponds to a patch $I_i^p$ processed by the multiple transformer blocks.
To integrate superpixels into the ViT framework, we need to build superpixel tokens from the existing patch tokens.
Following ~\cite{huang2023vision}, we adopt a more attention-like manner to compute the association map:
\begin{equation}
    Q^t = \text{Softmax} \left( \frac{\boldsymbol{T} {S^{t-1}}^{\rm T}}{\sqrt{d}} \right),
\end{equation}
where $d$ denotes the dimensionality $C$ of the token representations.
The superpixel token $S^t$ at iteration $t$ is computed as follows:
\begin{equation}
    S^{t} = (\hat{Q}^t)^{T}\boldsymbol{T},
\end{equation}
where $\hat{Q}^t$ is the column-normalized $Q^{t}$.
Similar to~\cite{jampani2018superpixel,huang2023vision}, to reduce computation burden, for each token, we restrict the corresponding local $3 \times 3$ surrounding superpixel tokens.
$S^{0}$ is obtained by performing an average pooling with grid size ${H_t\times W_{t}}$ on $\boldsymbol{T}$. The stride remains the same as the grid size.

\begin{figure*}[t]
    \includegraphics[width=\linewidth]{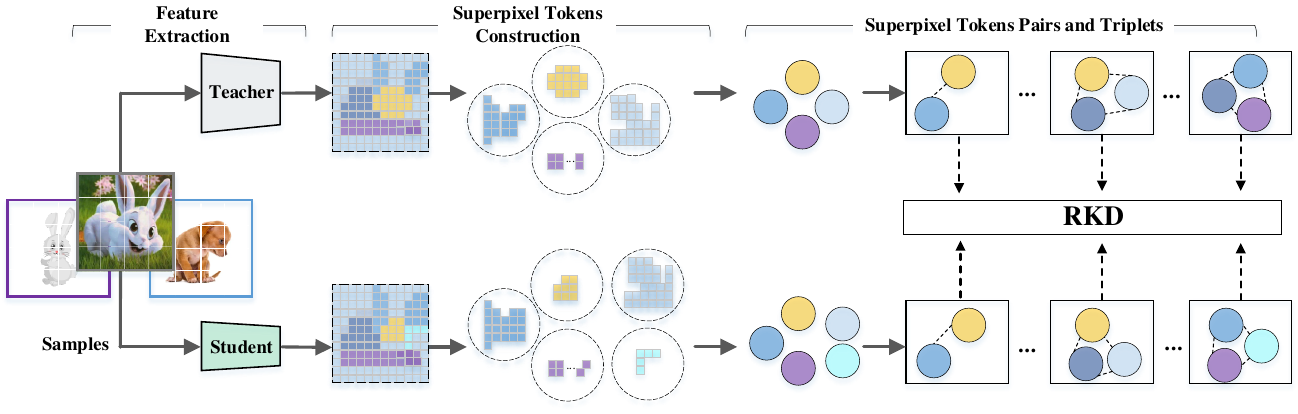}
    \caption{Illustration of the proposed SeRKD method, which mainly contains the mechanism of the feature extraction, the construction of semantic superpixel tokens, and the alignment of relation knowledge upon the superpixel tokens.}
    \label{fig:method}
\end{figure*}

\subsubsection{Building Superpixel Tokens on CNNs}
Given we only build the associations around its $9$ surrounding tokens due to the large computation burden of building the associations for all the tokens, we refine the superpixel tokens with a tiny learnable network $\Phi$.
For CNN features, which lack the structured patch tokens of Vision Transformers (ViTs), we propose a tokenization strategy to segment the CNN feature map into discrete tokens. Given the feature map $\boldsymbol{F}$, we define the tokenization operation as:
\begin{equation}
\mathbf{P} = \text{Tokenize}(\boldsymbol{F}, \Theta, H_p, W_p),
\end{equation}
where $\mathbf{P} \in \mathcal{R}^{L \times C}$ is the matrix of tokens, $L = \frac{H \times W}{H_p \times W_p}$ is the number of tokens, and $H_p$, $W_p$ are the dimensions of the tokenizing window. The parameter set $\Theta$ defines the tokenizer's rules, which can be parameter-free (e.g., max-pooling, average-pooling) or parameterized (e.g., strided-convolution).
This tokenization method enables the construction of superpixel tokens from CNN features, facilitating enhanced image representation analogous to ViTs.

\subsection{Relation Knowledge Distillation on Superpixel Tokens}
\label{sec:RKD_superpixel}

Traditionally, the application of Relation-based Knowledge Distillation (RKD) in Vision Transformers (ViTs) has not been explicitly explored, particularly with respect to the direct use of raw token representations. Recognizing this gap, we initially experimented with RKD on raw ViT tokens.
However, these preliminary attempts highlighted a significant performance degradation compared with the traditional KD~\cite{hinton2015distilling} method, suggesting the inadequacy of direct RKD application on raw tokens within the ViT framework.
To address this challenge, we introduce a novel approach that firstly constructs semantic superpixel tokens as described in ~\ref{sec:building_seman}. Subsequently, RKD is applied to these superpixel tokens, thereby aligning the distillation process more closely with the inherent structure and semantic layout of the data.

\subsubsection{Semantic Enhancement through Superpixel Tokens}
In this paper, we pivot from conventional instance-based relational learning to a more semantically grounded methodology. Utilizing superpixel tokens, detailed in Section \ref{sec:building_seman}, we enrich the visual tokens with semantic coherence. This semantic enrichment forms the foundation for our adapted RKD process.
We use $\mathbf{S}$ to represent the set of superpixel tokens derived from the visual tokens $\mathbf{T}$ via the superpixel clustering method described in Section \ref{sec:building_seman}.

\subsubsection{Semantics-based RKD Loss}
Building on these semantically enriched tokens, we recalibrate the RKD loss to operate in this new semantic token space. The distance-wise and angle-wise components of RKD loss are redefined to reflect the semantic relationships within the superpixel tokens. The revised RKD losses for pairs and triplets of superpixel tokens are formulated as follows:

\begin{align}
    \mathcal{L}_{\text{RD}}^{\text{SP}} = \frac{1}{\nu'} \sum_{i,j} l_{\delta}(\psi_{\text{D}}(s_i, s_j), \psi_{\text{D}}(s_i', s_j')),
\end{align}
\begin{align}
    \mathcal{L}_{\text{RA}}^{\text{SP}} = \sum_{i,j,k} l_{\delta}(\psi_{\text{A}}(s_i, s_j, s_k), \psi_{\text{A}}(s_i', s_j', s_k')),
\end{align}
where $s_i, s_j, s_k$ represent the superpixel tokens for the student model, $s_i', s_j', s_k'$ are for the teacher model, and $\nu'$ is a normalization factor for the superpixel-based RKD.

\subsubsection{Semantic-focused RKD}
Our method emphasizes the crucial role of semantics in building relational knowledge. By focusing on semantic construction prior to distillation, the model internalizes more meaningful and contextually relevant relationships.

Similar to RKD~\cite{Park_2019_CVPR} adopted in classification task, Knowledge distillation loss $\mathcal{L}_{\text{KD}}$~\cite{hinton2015distilling} and cross-entropy classification loss $\mathcal{L}_{\text{cls}}$ are also used, resulting the final distillation loss $\mathcal{L}_{\text{dis}}$:
\begin{align}
\mathcal{L}_{\text{dis}} = \mathcal{L}_\mathrm{cls} + \lambda_{K}\mathcal{L}_\mathrm{KD} + \lambda_{F}\mathcal{L}_{F} + \lambda_{D} \mathcal{L}^{\text{SP}}_{\text{RD}} + \lambda_{A}\mathcal{L}^{\text{SP}}_{\text{RA}},
\end{align}
where $\lambda_{K}$, $\lambda_{F}$, $\lambda_{D}$ and $\lambda_{A}$ are the hyper-parameters to balance the losses.
$\mathcal{L}_{F}$ is defined as the $\mathcal{L}_2$ loss between student feature $F_{s}$ and teacher feature $F_{t}$.
$\mathcal{L}_\mathrm{KD}$ is derived from ~\cite{hinton2015distilling} formulated as: 
\begin{align}
\label{eq:dark_knowledge_distillation}
\sum_{I_i} \mathrm{KL}\Big(\text{softmax}\big(\frac{l_t(I_i)}{\tau}\big), \text{softmax}\big(\frac{l_s(I_i)}{\tau}\big)\Big),
\end{align}
where $l_t(I_i)$/$l_s(I_i)$ denotes the logits (pre-softmax activation outputs) of the teacher/student model for a given input image $I_i$. $\tau$ is used to control the smoothness of the output probability distribution obtained from the softmax function applied to the logits.

\section{Experimental Analysis}
\label{sec:experimental_analysis}
This section presents a comprehensive experimental evaluation of our proposed Semantics-based Relation Knowledge Distillation (SeRKD). We primarily focus on the ImageNet-1k dataset\cite{imagenet} to assess SeRKD's effectiveness. Additionally, we extend our experiments to transfer learning tasks, demonstrating SeRKD's robust generalization capabilities. Ablation studies and detailed visualizations are also provided for a deeper understanding of the method.
$T=1$, $\lambda_{K}=1$, $\lambda_{F}=1$, $\lambda_{D}=0.5$ and $\lambda_{A}=1$ by default.

\subsection{Implementation Details}
\label{subsec:impl}
Our experiments utilize an MAE-base~\cite{he2022masked} as the teacher, attaining a top-1 accuracy of $83.6\%$ on ImageNet-1k.
Each model undergoes training for $300$ epochs using a batch size of $1024$, trained on $8$ NVIDIA V100 GPUs, \ie, $128$ images per GPU.
We employ the AdamW optimizer with an initial learning rate of $0.001$, employing a cosine decay learning rate schedule.
The weight decay parameter is set to $0.05$.
For input images, we use a resolution of $224\times224$. Our data augmentation strategy includes Mixup, Cutmix, and RandAugment.
Following DeiT~\cite{deit}, an additional distillation token is employed in a similar fashion to the class token for the student network. For classification on ImageNet, the total number of tokens is $14 \times 14 + 1 + 1 = 198$. Apart from the conventional classification head applied to the class token, a distillation head is also applied to the distillation token. We calculate the average value of the logits produced by the classification head and the distillation head to obtain the final prediction. For $\mathcal{L}_{KD}$, it is applied to the logits produced by the distillation head.

We use the visual tokens (with the class token and distillation token removed) from the final transformer block to perform superpixel clustering and relation knowledge distillation. To this end, $\mathcal{L}_{F}$ is applied to these visual tokens. These visual tokens are further used to perform superpixel clustering for relation distillation. $H_t$ and $W_t$ are both set to $2$, indicating that the initial superpixels $S^{0}$ are obtained by performing average pooling on the visual tokens. Therefore, the superpixel size is $14/2 \times 14/2 = 7 \times 7$. We set the iteration $T=1$ by default, as we find that more iterations do not yield better performance.

We train variants of our SeRKD model: SeRKD-Ti (SeRKD-Tiny), SeRKD-S (SeRKD-Small), the student architecture is the same as DeiT~\cite{deit}.

\subsection{Implementation and Complexity of RKD Loss on Superpixel Tokens}
Here are the implementations of the angle-wise and distance-wise loss functions for relation-based knowledge distillation in Vision Transformers.
\texttt{RkdDistance\_token} and \texttt{RkdAngle\_token} are the functions of $\mathcal{L}^{SP}_{RD}$ and $\mathcal{L}^{SP}_{RA}$, respectively.

\begin{lstlisting}[language=Python, caption=Angle-wise Loss Calculation]
def extended_rkdangle(vit_feat):
    B, L, C = vit_feat.shape
    # Compute pairwise differences
    diff = vit_feat.unsqueeze(2) - vit_feat.unsqueeze(1)  # Shape: [B, L, L, C]
    norm_diff = F.normalize(diff, p=2, dim=3)  # Shape: [B, L, L, C]
    # Reshape for batch matrix multiplication
    norm_diff_flat = norm_diff.view(B * L, L, C)
    # Perform batch matrix multiplication to get the angle potentials
    angle_flat = torch.bmm(norm_diff_flat, norm_diff_flat.transpose(1, 2))
    # Reshape back to original dimensions
    angle = angle_flat.view(B, L, L, L)
    return angle
class RKdAngle_token(nn.Module):
    def forward(self, student, teacher):
        with torch.no_grad():
            t_angle = extended_rkdangle(teacher).view(-1)
        s_angle = extended_rkdangle(student).view(-1)
        loss = F.smooth_l1_loss(s_angle, t_angle, reduction='mean')
        return loss

def batch_pdist(e, squared=False, eps=1e-12):
    e_square = e.pow(2).sum(dim=2)
    prod = torch.einsum('bik,bjk->bij', e, e)  # Pairwise dot products
    res = (e_square.unsqueeze(2) + e_square.unsqueeze(1) - 2 * prod).clamp(min=eps)
    if not squared:
        res = res.sqrt().clone()
    # Set the diagonal to zero across all batches
    res[:, torch.arange(res.size(1)), torch.arange(res.size(2))] = 0
    return res
class RkdDistance_token(nn.Module):
    def forward(self, student, teacher):
        L = student.shape[1]
        with torch.no_grad():
            t_d = batch_pdist(teacher, squared=False)  # Shape: [B, L, L]
            mean_td = t_d.sum(dim=(1, 2), keepdim=True) / (L * (L - 1))  # Normalize
            t_d = t_d / mean_td
        d = batch_pdist(student, squared=False)  # Shape: [B, L, L]
        mean_d = d.sum(dim=(1, 2), keepdim=True) / (L * (L - 1))  # Normalize
        d = d / mean_d
        loss = F.smooth_l1_loss(d, t_d, reduction='mean')
        return loss

loss_distill_angle = RKdAngle_token()
loss_distill_rkd_dist = RkdDistance_token()   
loss_rkd_angle = args.angle_w * loss_distill_angle(stu_super_tokens, tec_super_tokens.detach())
loss_rkd_dist = args.dist_w * loss_distill_rkd_dist(stu_super_tokens, tec_super_tokens.detach())
\end{lstlisting}

Given an input superpixel tensor with shape $(B, L, C)$, where $B$ is the batch size, $L$ is the number of superpixel tokens, and $C$ is the feature dimension, the complexity analysis is as follows:

\paragraph{Complexity Analysis of \texttt{RKdAngle\_token}}~{}

\textbf{Time Complexity:}

\begin{itemize}
    \item Pairwise Difference Computation: $\mathcal{O}(B L^2 C)$
    \item Normalization: $\mathcal{O}(B L^2 C)$
    \item Batch Matrix Multiplication: $\mathcal{O}(B L^2 C)$
    \item Reshaping: $\mathcal{O}(1)$ (negligible)
\end{itemize}

Overall, the time complexity is $\mathcal{O}(B L^2 C)$.

\textbf{Space Complexity:}

\begin{itemize}
    \item Intermediate Tensors: $\mathcal{O}(B L^2 C)$ each
    \item Angle Tensor: $\mathcal{O}(B L^3)$
\end{itemize}

Overall, the space complexity is $\mathcal{O}(B L^3)$.

\paragraph{Complexity Analysis of \texttt{RkdDistance\_token}}~{}

\textbf{Time Complexity:}

\begin{itemize}
    \item Pairwise Distance Computation: $\mathcal{O}(B L^2 C)$
    \item Normalization: $\mathcal{O}(B L^2)$
    \item Loss Calculation: $\mathcal{O}(B L^2)$
\end{itemize}

Overall, the time complexity is $\mathcal{O}(B L^2 C)$.

\textbf{Space Complexity:}

\begin{itemize}
    \item Pairwise Distance Tensor: $\mathcal{O}(B L^2)$
\end{itemize}

Overall, the space complexity is $\mathcal{O}(B L^2)$.

\textbf{Remark:} From the analysis above, it is evident that the primary space complexity arises from the \texttt{RKdAngle\_token} function. For instance, with float16 precision, the memory consumption of \texttt{RKdAngle\_token} is approximately 0.95 GB for $B=128$, $L=49$, and $C=768$. However, when $L$ increases to $14 \times 14 = 196$, the memory usage escalates to 17.73 GB, leading to the risk of out-of-memory (OOM) errors. Therefore, for our setting, we recommend using $H_t = W_t = 2$ to ensure the number of superpixel tokens is $49$, making the computation and space complexity more manageable.

\subsection{Performance on ImageNet-1k}
We evaluate SeRKD on the ImageNet-1k\cite{imagenet} dataset, a prominent large-scale image classification benchmark with $1.28$ million training images and $50,000$ validation images across $1,000$ categories. ImageNet-1k offers a challenging environment to test the efficacy of our SeRKD approach.

\subsubsection{ImageNet classification on SeRKD-ViT}
The results in Table~\ref{tab:imagenet} highlight SeRKD's superior performance on ImageNet-1k.

\begin{table}[th]
    \centering
    \begin{small}
    \caption{\textbf{Results on ImageNet-1k}. * means that RegNets are optimized with similar optimization procedures as DeiT, serving as teachers for DeiT and our CSKD. ``val-real" represents the results of the ImageNet Real validation set.}
    \label{tab:imagenet}
    \begin{tabular}{lcc|c}
\multicolumn{1}{c|}{Method}    & \#params(M) & \begin{tabular}[c]{@{}c@{}}image\\ size\end{tabular} & \begin{tabular}[c]{@{}c@{}}val\\ \end{tabular} \\ \Xhline{3\arrayrulewidth}
\multicolumn{4}{c}{\textit{CNNs}} \\ 
\hline
\multicolumn{1}{c|}{ResNet18\cite{resnet}} & 12M & $224^2$ & 69.8 \\% & 77.3 \\
\multicolumn{1}{c|}{ResNet50\cite{resnet}} & 25M & $224^2$ & 76.2 \\% & 82.5 \\  
\multicolumn{1}{c|}{ResNet101\cite{resnet}} & 45M & $224^2$ & 77.4 \\% & 83.7 \\  
\multicolumn{1}{c|}{ResNet152\cite{resnet}} & 60M & $224^2$ & 78.3 \\% & 84.1 \\
\hline
\multicolumn{1}{c|}{RegNetY-4GF\cite{radosavovic2020designing}} & 21M & $224^2$ & 80.0 \\%& 86.4 \\  
\multicolumn{1}{c|}{RegNetY-8GF\cite{radosavovic2020designing}} & 39M & $224^2$ & 81.7 \\%& 87.4 \\ 
\multicolumn{1}{c|}{RegNetY-16GF\cite{radosavovic2020designing}} & 84M & $224^2$ & 82.9 \\%& 88.1 \\ 
\hline
\multicolumn{1}{c|}{EffiNet-B0\cite{tan2019efficientnet}} & 5M & $224^2$ & 77.1 \\%& 84.1 \\
\multicolumn{1}{c|}{EffiNet-B3\cite{tan2019efficientnet}} & 12M & $224^2$ & 81.6 \\%& 86.8 \\
% \multicolumn{1}{c|}{EffiNet-B6\cite{tan2019efficientnet}} & 43M & $224^2$ & 84.0 \\%& 88.8 \\
\hline
\multicolumn{4}{c}{\textit{ViTs}} \\ 
\hline
\multicolumn{1}{c|}{ViT-B/16\cite{vit}} & 86M & $384^2$ & 77.9 \\%& 83.6 \\
\multicolumn{1}{c|}{ViT-L/16\cite{vit}} & 307M & $384^2$ & 76.5 \\%& 82.2 \\
\hline
\multicolumn{1}{c|}{DeiT-Ti\cite{deit}} & 6M & $224^2$ & 74.5 \\%& 82.1 \\
\multicolumn{1}{c|}{DeiT-S\cite{deit}} & 22M & $224^2$ & 81.2 \\%& 86.8 \\
% \multicolumn{1}{c|}{DeiT-B\cite{deit}} & 87M & $224^2$ & 83.4 \\%& 88.3 \\
\hline
\multicolumn{1}{c|}{DearKD-Ti\cite{chen2022dearkd}} & 5M & $224^2$ & 74.8 \\%& - \\
\multicolumn{1}{c|}{DearKD-S\cite{chen2022dearkd}} & 22M & $224^2$ & 81.5 \\%& - \\
% \multicolumn{1}{c|}{DearKD-B\cite{chen2022dearkd}} & 86M & $224^2$ & 83.6 \\%& - \\
\hline
\multicolumn{1}{c|}{CSKD-Ti\cite{zhao2023cumulative}} & 6M & $224^2$ & 76.3 \\%& \textbf{83.6} \\
\multicolumn{1}{c|}{CSKD-S\cite{zhao2023cumulative}} & 22M & $224^2$ & 82.3 \\%& \textbf{87.9} \\
% \multicolumn{1}{c|}{CSKD-B} & 87M & $224^2$ & \textbf{83.8} \\%& \textbf{88.6} \\

\hline
\multicolumn{1}{c|}{\textbf{SeRKD-Ti}} & 6M & $224^2$ & \textbf{76.8} \\
\multicolumn{1}{c|}{\textbf{SeRKD-S}} & 22M & $224^2$ &   \textbf{82.5} \\

\bottomrule

    \end{tabular}
    \end{small}
\end{table}

\subsubsection{ImageNet Classification on CNN}
Implementing SeRKD on CNN architectures necessitates the tokenization of CNN features, adapting them for relational knowledge distillation. We investigate different tokenization methods to determine the optimal approach for distilling knowledge from ResNet-101 to ResNet-18.

We adopt the feature maps of the first three stages for knowledge distillation. For the $1$-st, $2$-nd, and $3$-rd stages, we perform pooling or stride convolution with stride=$4, 2, 1$, respectively.
The number of image tokens of each stage is kept at $14\times 14=196$.
Both student and teacher models share the same tokenizer setup to ensure compatible feature representation. Notably, for parameterized tokenizers like strided convolutions, teacher gradients are detached to ease the learning of the student and the tokenizer.

\begin{table}[th]
    \centering
    \begin{small}
    \caption{Performance of different tokenization methods for student ResNet-18 with teacher ResNet-101 on ImageNet-1k.}
    \label{tab:imagenet}
    \begin{tabular}{l|c|c}
    \multicolumn{1}{c|}{Method}    & Tokenizer & \begin{tabular}[c]{@{}c@{}}Top-1 Accuracy (\%)\end{tabular} \\ \Xhline{3\arrayrulewidth}
    \multicolumn{1}{c|}{KD \cite{hinton2015distilling}} & - &  71.09 \\
    \hline
    \multicolumn{1}{c|}{RKD \cite{Park_2019_CVPR}} & - &  71.82 \\
    \hline
    \multirow{3}{*}{SeRKD} & Max Pooling &  70.61 \\ \cline{2-3} 
                           & Average Pooling &  72.22 \\ \cline{2-3}
                           & Strided Convolutions &  \textbf{72.25} \\ 
    \bottomrule
    \end{tabular}
    \end{small}
\end{table}

Strided convolutions and average pooling deliver the top two performances with only marginal differences, underscoring their effectiveness in preserving spatial hierarchies crucial for detailed feature transfer. However, max pooling shows slightly reduced efficacy, likely due to its potential for detail loss.
When employing average pooling or strided convolutions as tokenizers, our SeRKD method notably surpasses traditional KD and RKD, achieving significant performance gains.

\subsection{Transfer Learning to Downstream Datasets}
\label{sec:transfer_learning}

We conduct experiments to evaluate the effectiveness of our proposed method in the context of transfer learning to various downstream datasets. The datasets used for these experiments include CIFAR10 \cite{cifar}, CIFAR100 \cite{cifar} and Cars \cite{krause20133d}. Detailed information about each dataset is presented in Table~\ref{tab:transfer_dataset}.

Our experimental results are summarized in Table~\ref{tab:transfer}. As can be observed, our proposed method, denoted as SeRKD, achieves significantly superior transfer learning performances compared to the DeiT and DearKD approaches across all considered downstream datasets. Specifically, our method SeRKD-S surpasses CSKD-S by $0.1\%$, $1.0\%$, and $0.4\%$ over CIFAR10, CIFAR100, and Cars.

\begin{table}[ht]
    \centering
    \begin{small}
    \caption{Downstream dataset information for transfer learning.}
    \label{tab:transfer_dataset}
    \begin{tabular}{l|ccc}
        dataset & training size & val size & \#classes  \\
        \Xhline{3\arrayrulewidth}
        CIFAR10 & 50,000 & 10,000 & 10 \\
        CIFAR100 & 50,000 & 10,000 & 100 \\
        Stanford Cars & 8,144 & 8,041 & 196 \\
    \end{tabular}

    \end{small}
\end{table}

\begin{table}[ht]
\centering
\begin{small}
\caption{\textbf{Transfer performance} in downstream tasks. * represents that the results are based on our implementation.}
\label{tab:transfer}
\begin{tabular}{l|ccc}
  & CIFAR10 & CIFAR100 & Cars   \\
 \Xhline{3\arrayrulewidth}
 DeiT-Ti* & 98.1 & 86.1 & 92.1 \\
 DearKD-Ti & 97.5 & 85.7 &  89.0 \\
 CSKD-Ti & 98.5 & 87.0 & 93.1 \\
 SeRKD-Ti & \textbf{98.6} & \textbf{87.8} & \textbf{93.6}  \\
 \hline
 DeiT-S* & 98.7 & 89.2 & 91.7 \\
 DearKD-S & 98.4 & 89.3 & 91.3 \\
 CSKD-S & 99.1 & 90.3 & 93.7 \\
 SeRKD-S & \textbf{99.2} & \textbf{91.3} & \textbf{94.1} \\
 % \hline
 % DeiT-B* & 99.1 & 91.3 & 92.9 & 82.1 \\
 % DearKD-B & 99.2 & 91.1 & 92.7 & - \\
 % CSKD-B & \textbf{99.3} & \textbf{91.4} & \textbf{94.0} & \textbf{82.5} \\
\end{tabular}

\end{small}
\end{table}

\section{Ablation Study}
\label{sec:ablation}

\subsection{Ablations on the Key Components of SeRKD}
To understand the contributions of different components of our proposed SeRKD framework, we conduct a series of ablation experiments.
The results in Table~\ref{tab:performance_of_conponets} demonstrate the effectiveness of each component within the SeRKD framework. Specifically, we observe the following:
The results in Table~\ref{tab:performance_of_conponets} demonstrate the effectiveness of each component within the SeRKD framework. 
Knowledge Distillation (KD)~\cite{hinton2015distilling} method achieves an accuracy of $74.1\%$. By incorporating the feature-based loss $\mathcal{L}_F$, FitNet~\cite{romero2014fitnets} shows a marginal improvement with an accuracy of $74.2\%$.
Secondly, introducing superpixel-based clustering along with feature-based and distance-wise relational losses ($\mathcal{L}_F$ and $\mathcal{L}^{SP}_{RD}$), SeRKD-Ti$\dagger$ significantly boosts the accuracy to $76.5\%$. 
When angle-wise relational loss ($\mathcal{L}^{SP}_{RA}$) is used in conjunction with superpixel clustering and distance-wise relational loss, the performance slightly decreases to $76.2\%$, highlighting the critical balance between these components.
Finally, the full SeRKD-Ti model, integrating superpixel clustering with all relational and feature-based losses ($\mathcal{L}_F$, $\mathcal{L}^{SP}_{RD}$, and $\mathcal{L}^{SP}_{RA}$), achieves the highest accuracy of $76.8\%$, underscoring the complementary nature of these components.
These ablation studies illustrate the importance of each component in enhancing the distillation process, validating our design choices for the SeRKD framework.

\begin{table}[ht]
\centering
  \centering
  \renewcommand\arraystretch{1.8}
  \setlength\tabcolsep{3mm}
  \scriptsize
  \caption{Ablations on the key components of SeRKD.}
  \label{tab:performance_of_conponets}
  \centering
  \begin{tabular}{l|c|c|c|c|c}
    \toprule[1pt]
    Methods & Clustering  & $\mathcal{L}_F$ & $\mathcal{L}^{SP}_{RD}$ & $\mathcal{L}^{SP}_{RA}$ & Acc (\%) \\
    \hline 
    KD~\cite{hinton2015distilling} & {} & {} & {} & {} & 74.1 \\
    \hline
    Fitnet~\cite{romero2014fitnets} & {}  & $\checkmark$ & {}  & {} & 74.2 \\
    \hline
    SeRKD-Ti$\dagger$  & Superpixel  & $\checkmark$ & $\checkmark$ & {} & 76.5 \\
    \hline
    SeRKD-Ti$\ddagger$  & Superpixel  & {} & $\checkmark$ & $\checkmark$ & 76.2 \\
    \hline
    SeRKD-Ti & Superpixel  & $\checkmark$ & $\checkmark$ & $\checkmark$ & \textbf{76.8} \\
    \bottomrule[1pt]
  \end{tabular}

\end{table}

\subsection{Ablation on Clustering Methods}
\label{sec:token_merge}
In order to perform relation knowledge distillation on image tokens effectively, it is essential to first construct meaningful semantic representations. In this subsection, we explore three approaches for building semantics: direct-based, pooling-based (MaxPooling and AvgPooling), and superpixel-based.

The direct-based approach involves applying relation knowledge distillation directly on the image tokens without any prior semantic construction. Pooling-based methods aim to create semantics by aggregating tokens through pooling operations. The superpixel-based approach, which is ultimately adopted in our method, leverages the inherent semantic layout of the image to guide the token merging process.
To establish a baseline for comparison, we remove the superpixel-based RKD loss from the training objective by setting $\lambda_D=\lambda_A=0$.

The results presented in Table~\ref{tab:token_merge} yield several key insights. When applying relation knowledge distillation directly on the image tokens, we observe a substantial performance degradation of $1.2\%$ compared to the baseline setting. This significant drop in performance can be attributed to the large number of tokens ($196$) and the lack of essential semantic relations within individual tokens, hindering effective distillation.
Employing pooling-based token merging methods leads to improved performance compared to the direct-based approach. However, the results still fall short of the baseline. This suggests that hard pooling methods alone are insufficient for capturing the complex semantic relationships inherent in image tokens.

In contrast, the superpixel-based token merging method, which respects the semantic layout of the image, surpasses the baseline approach, achieving a top-1 accuracy of $77.4\%$. This represents a notable improvement of $2.3\%$ over the baseline, highlighting the effectiveness of leveraging semantic information in the token clustering process.

\begin{figure*}[th]
    \centering
    \includegraphics[width=0.8\linewidth]{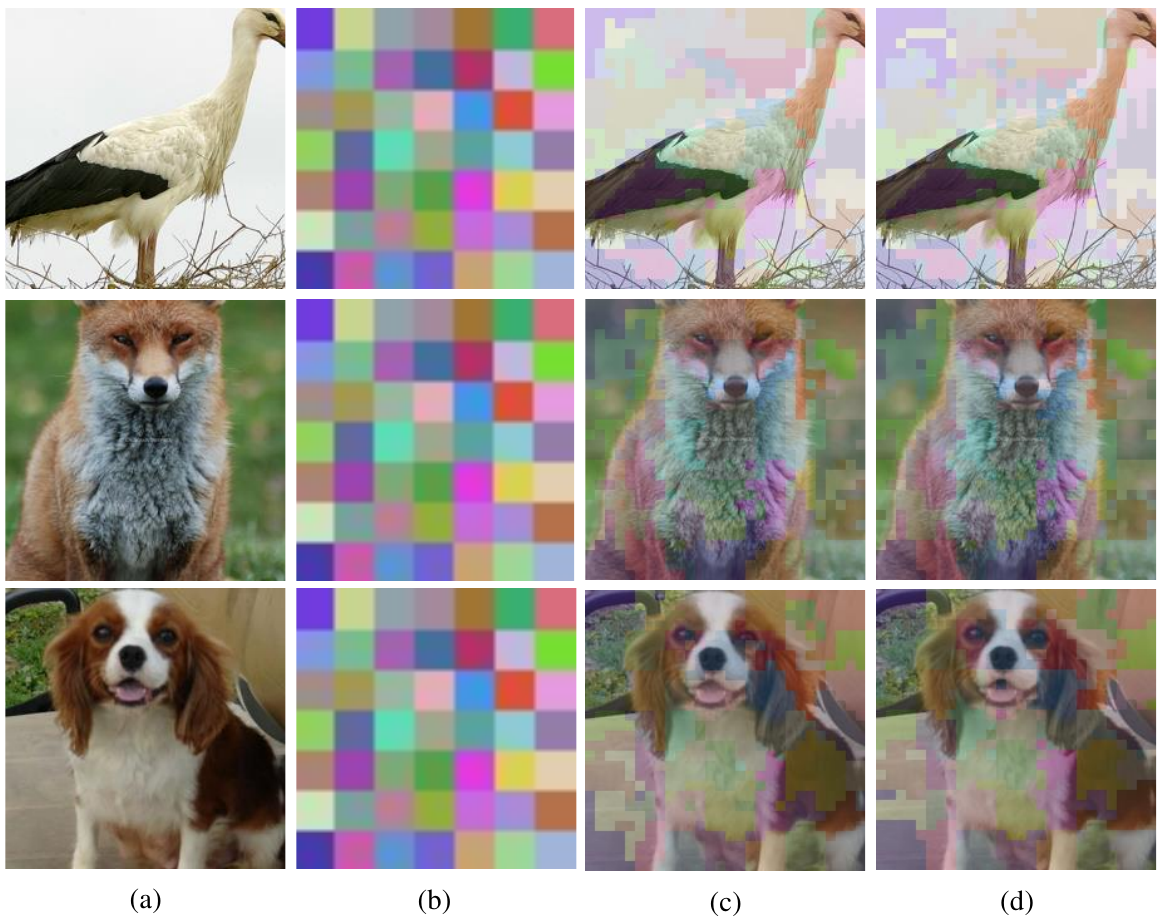}
    \caption{Visualization of learned superpixel tokens in the SeRKD-S distillation setting. (a) is the input image, (b) is the superpixel map, (c) shows the superpixel tokens of the teacher, and (d) shows the learned superpixel tokens of the student.}
    \label{fig:superpixel}
\end{figure*}

\begin{table}[ht]
\centering
\begin{small}
\caption{\textbf{Performance of SeRKD-Ti with different token merging methods.} For the Direct-based method, we reduce the batch size to $32$ per GPU to avoid OOM.}
\begin{tabular}{l|c|c}
\toprule
\textbf{Methods} & Top-1 ($\%$) & Top-5 ($\%$) \\
\midrule
Baseline & 74.2 & 92.2 \\
Direct-based & 73.9 & 92.0 \\
MaxPooling-based & 73.3 & 91.5 \\
AvgPooling-based & 73.8 & 91.7 \\
\textbf{Superpixel-based} & \textbf{76.8} & \textbf{93.4} \\
\bottomrule
\end{tabular}

\label{tab:token_merge}
\end{small}
\end{table}

\subsection{Visualization of Learned Superpixels}

Figure~\ref{fig:superpixel} visualizes the superpixel tokens of the SeRKD-S distillation setting. In Figure~\ref{fig:superpixel}(c), the superpixel tokens of the teacher MAE-base are shown, and Figure~\ref{fig:superpixel}(d) displays the learned superpixel tokens of the student SeRKD-S. It is evident that these two types of superpixels share similar semantic regions, indicating that our RKD loss on these superpixel tokens effectively enhances the alignment of the semantic layout between the teacher and the student.

\subsection{Ablation on Superpixel-RKD Loss}
\label{sec:rkd_super_params}
To investigate the impact of the Superpixel-RKD loss on the performance of SeRKD, we conduct an ablation study by varying the hyper-parameters $\lambda_{D}$ and $\lambda_{A}$. Table~\ref{tab:rkd_super_params} presents the results of this study.
Initially, we perform a search over different values of $\lambda_{K}$ to establish a baseline model. Based on these preliminary experiments, we set $\lambda_{K}=1$ for the baseline.
Building upon this baseline, we explore the performance of SeRKD under various settings of $\lambda_{D}$ and $\lambda_{A}$. The results, summarized in Table~\ref{tab:rkd_super_params}, demonstrate the impact of these hyper-parameters on the model's performance.
After careful consideration of the results, we determine the optimal values of $\lambda_{D}=0.5$ and $\lambda_{A}=1$ for all our SeRKD training models. With these settings, SeRKD-Ti and SeRKD-S achieve top-1 accuracies of $76.8\%$ and $82.5\%$, respectively, on the ImageNet dataset.

\begin{table}[ht]
\renewcommand\arraystretch{1.2}
\setlength\tabcolsep{3mm}
\caption{Performance of SeRKD with different $\lambda_{K}$, $\lambda_{D}$, and $\lambda_{A}$ values on ImageNet.}
\centering
\begin{tabular}{c|c|c|c|c|c}
\toprule[1pt]
$\lambda_{K}$ & $\lambda_{F}$ & $\lambda_{D}$ & $\lambda_{A}$ & SeRKD-Ti & SeRKD-S \\
\hline
0.1 & 0 & 0 & 0 & 73.0 & 80.7 \\
\hline
0.5 & 0 & 0 & 0 & 73.8 & 81.0 \\
\hline
1 & 0 & 0 & 0 & 74.1 & 81.2 \\
\hline
2 & 0 & 0 & 0 & 74.0 & 81.0 \\
\hline
1 & 1 & 0 & 0 & 74.2 & 81.3 \\
\hline
1 & 2 & 0 & 0 & 74.2 & 81.2 \\
\hline
1 & 1 & 0.1 & 0.1 & 76.1 & 81.9 \\
\hline
1 & 1 & 0.5 & 0.5 & 76.6 & 82.4 \\
\hline
1 & 1 & 1.0 & 1.0 & 76.2 & 82.0 \\
\hline
1 & 1 & 0.5 & 0.1 & 76.4 & 82.2 \\
\hline
1 & 1 & 0.5 & 1.0 & \textbf{76.8} & \textbf{82.5} \\
\hline
1 & 1 & 0.5 & 2.0 & 76.7 & \textbf{82.5} \\
\bottomrule[1pt]
\end{tabular}
\label{tab:rkd_super_params}
\end{table}

\subsection{Ablation on Grid Size}

We investigate the impact of grid sizes ($H_t \times W_t$) on the performance of SeRKD-Ti and SeRKD-S models. Table~\ref{tab:grid_size} presents the top-1 accuracy results with grid sizes of $1\times 1$, $2\times 2$, and $3\times 3$. For the $1\times 1$ grid, the batch size was reduced to $32$ per GPU to avoid OOM issues.

For the smallest grid size ($1\times 1$), the performance is lower (73.9\% for SeRKD-Ti and 79.7\% for SeRKD-S), likely due to the limited context each superpixel captures and the increased complexity, resulting in less effective relational knowledge distillation.
The $2\times 2$ grid size yields the highest accuracy (76.6\% for SeRKD-Ti and 82.4\% for SeRKD-S), suggesting it strikes a balance between capturing sufficient local context and maintaining computational efficiency.
With the largest grid size ($3\times 3$), the performance decreases again (74.5\% for SeRKD-Ti and 80.3\% for SeRKD-S). This may be because the superpixels contain too large a context; in this case, a superpixel token aggregates a $48\times 48$ region for a $224\times 224$ image, making the relational knowledge less specific and effective.

In conclusion, the $2\times 2$ grid size is optimal for the SeRKD method, providing a good trade-off between capturing local and global information and maintaining computational feasibility.

\begin{table}[th]
\centering
\begin{small}
\caption{\textbf{Performance of SeRKD-Ti/S with different grid size.} For grid size $1\times 1$, we reduce the batch size to $32$ per GPU to avoid OOM.}
\begin{tabular}{l|c|c}
\toprule
\textbf{Methods} & $H_t \times W_t$ & Top-1 ($\%$) \\
\midrule
SeRKD-Ti & $1\times 1$ & 73.9 \\
SeRKD-S & $1\times 1$ & 79.7 \\
SeRKD-Ti & $2\times 2$ & \textbf{76.6} \\
SeRKD-S & $2\times 2$ & \textbf{82.4} \\
SeRKD-Ti & $3\times 3$ & 74.5 \\
SeRKD-S & $3\times 3$ & 80.3 \\
\bottomrule
\end{tabular}

\label{tab:grid_size}
\end{small}
\end{table}

\subsection{Ablation on Iteration Times $T$}

We conducted an ablation study to investigate the impact of the iteration count $T$ on the performance of our SeRKD models. We varied the number of iterations, specifically set at $1$, $2$, and $3$, to understand its effect on the model's top-1 accuracy.
The results, summarized in Table \ref{tab:token_merge}, demonstrate the top-1 accuracy achieved by the SeRKD-Ti and SeRKD-S models at different iteration times. It is observed that
for both Increasing SeRKD-Ti and SeRKD-S, the models perform best at $T=1$, A further increase in iteration count marginal decreases the performances.
The observed trend suggests that a higher number of iterations $T$ does not necessarily contribute to better performance. This could be attributed to the potential over-smoothing of features, where repeated aggregation may dilute distinctive features that are crucial for accurate classification.

\begin{table}[th]
\centering
\begin{small}
\caption{\textbf{Performance of SeRKD-Ti/S with different iteration times $T$.}}
\begin{tabular}{l|c|c}
\toprule
\textbf{Methods} & $T$ & Top-1 ($\%$) \\
\midrule
SeRKD-Ti & 1 & \textbf{76.8} \\
SeRKD-S & 1 & \textbf{82.5} \\
SeRKD-Ti & 2 & 76.6 \\
SeRKD-S & 2 & 82.4 \\
SeRKD-Ti & 3 & 76.2 \\
SeRKD-S & 3 & 82.1 \\
\bottomrule
\end{tabular}

\label{tab:token_iteration}
\end{small}
\end{table}

\section{Conclusion}
\label{sec:conclusion}
This paper introduced Semantics-based Relation Knowledge Distillation (SeRKD), a novel approach to knowledge distillation that leverages superpixels for semantic extraction and relation-based knowledge transfer. Our method demonstrated superior performance on benchmark datasets, particularly in the context of Vision Transformers (ViTs). 
SeRKD transcends traditional instance-level distillation techniques by incorporating semantic relationships, leading to enhanced model performance and generalization. This advancement not only offers superior efficiency in model compression but also opens new avenues for more nuanced and contextually rich knowledge transfer in various machine learning applications.

% \section*{Acknowledgments}
% TODO

\bibliographystyle{IEEEtran}
\bibliography{IEEEabrv,main}

% \bibliographystyle{IEEEtran}

% \end{thebibliography}

\newpage

\begin{IEEEbiography}[{\includegraphics[width=1in,height=1.25in,clip,keepaspectratio]{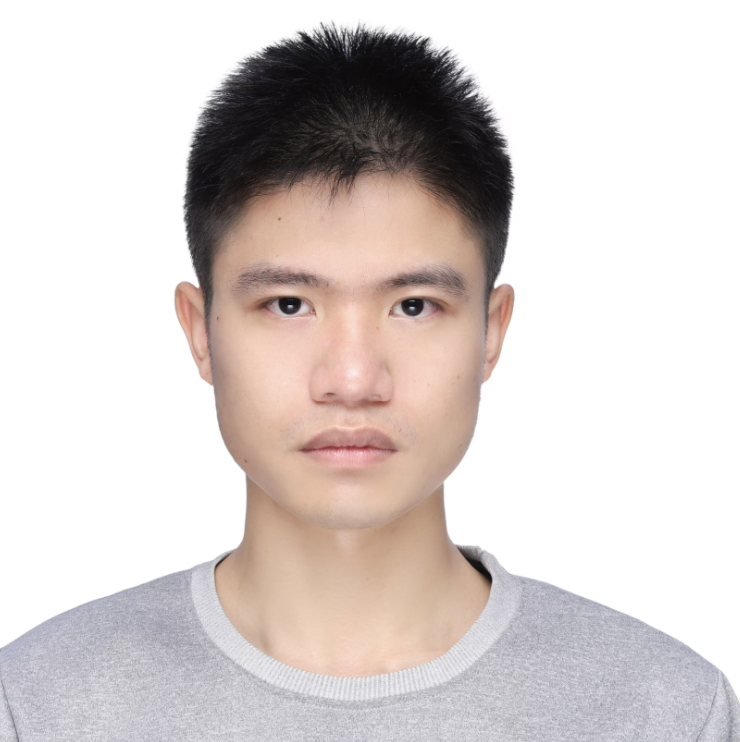}}]{Zhaoyi Yan}  received the Ph.D. degree in Computer Science from Harbin Institute of Technology, China, in 2021. His research interests include deep learning, image inpainting, crowd counting, knowledge distillation and Large Language Models. He has published more than 10 papers in conferences and journal including CVPR, ICCV, ECCV, AAAI, TNNLS and TCSVT.
\end{IEEEbiography}

\begin{IEEEbiography}[{\includegraphics[width=1in,height=1.25in,clip,keepaspectratio]{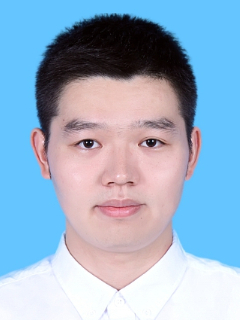}}]{Kangjun Liu} received the Ph.D. degree in information and communication engineering from South China University of Technology, China, in 2023. He is a postdoctoral researcher in Peng Cheng Laboratory, Shenzhen, China. His research interests include computer vision, representation learning and pattern recognition. He has published several papers in conferences and journals including TIP and PR.
\end{IEEEbiography}

\begin{IEEEbiography}[{\includegraphics[width=1in,height=1.25in,clip,keepaspectratio]{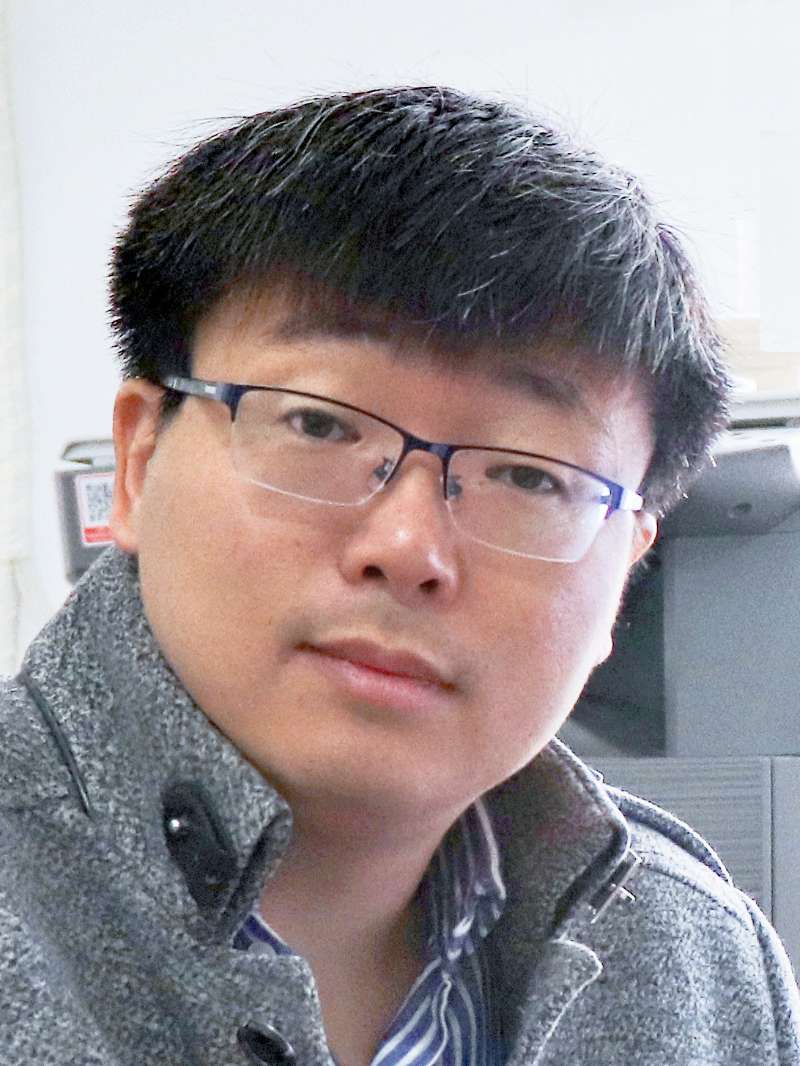}}] {Qixiang Ye} (M'10-SM'15) received the B.S. and M.S. degrees in mechanical and electrical engineering from Harbin Institute of Technology, China, in 1999 and 2001, respectively, and the Ph.D. degree from the Institute of Computing Technology, Chinese Academy of Sciences in 2006. He has been a professor with the University of Chinese Academy of Sciences since 2009, and was a visiting assistant professor with the Institute of Advanced Computer Studies (UMIACS), University of Maryland, College Park until 2013. His research interests include image processing, object detection and machine learning. He has published more than 100 papers in refereed conferences and journals including IEEE CVPR, ICCV, ECCV and TPAMI, TIP and TCSVT. He is on the editorial board of IEEE Transactions on Circuit and Systems on Video Technology. 
\end{IEEEbiography}

\vfill

\end{document}